\documentclass[11pt]{article}
\usepackage{acl2016}
\usepackage{times}
\usepackage{latexsym}
\usepackage{url}
\usepackage{algorithm}
\usepackage{algpseudocode}
\usepackage{pifont}
\usepackage{times}
\usepackage{url}
\usepackage{multirow}
\usepackage{amssymb}
\usepackage{amsthm}
\usepackage{amsmath}
\usepackage{verbatim}
\usepackage{tikz-qtree}
\usepackage{tikz}

\algrenewcomment[1]{\(\triangleright\) #1}
\algnewcommand{\LineComment}[1]{\State \(\triangleright\) #1}

\newenvironment{itemizesquish}{\begin{list}{\labelitemi}{\setlength{\itemsep}{-0.2em}\setlength{\labelwidth}{0.5em}\setlength
{\leftmargin}{\labelwidth}\addtolength{\leftmargin}{\labelsep}}}{\end{list}}

\newcommand{\ignore}[1]{}
\newcommand{\Sref}[1]{\S\ref{#1}}
\newcommand{\fref}[1]{figure~\ref{#1}}
\newcommand{\Fref}[1]{Figure~\ref{#1}}
\newcommand{\tref}[1]{table~\ref{#1}}

\newcommand{\Aref}[1]{Algorithm~\ref{#1}}
\usepackage[T1]{fontenc}

\aclfinalcopy 
\def\aclpaperid{697} 

\title{Learning the Curriculum with Bayesian Optimization \\for Task-Specific Word Representation Learning}
 \author{Yulia Tsvetkov$^{\spadesuit}$ ~ Manaal Faruqui$^{\spadesuit}$ ~ Wang Ling$^{\clubsuit}$ ~ Brian MacWhinney$^{\spadesuit}$ ~ Chris Dyer$^{\clubsuit\spadesuit}$ \\
 $^\spadesuit$Carnegie Mellon University ~~ $^\clubsuit$Google DeepMind  \\
{ \tt $\{$ytsvetko,mfaruqui,cdyer\}@cs.cmu.edu,} \\{\tt lingwang@google.com, macw@cmu.edu}}

\date{}

\begin{document}

\maketitle

\begin{abstract}
We use Bayesian optimization to learn curricula for word representation learning, optimizing performance on downstream tasks that depend on the learned representations as features. The curricula are modeled by a linear ranking function which is the scalar product of a learned weight vector and an engineered feature vector that characterizes the different aspects of the complexity of each instance in the training corpus. We show that learning the curriculum improves performance on a variety of downstream tasks over random orders and in comparison to the natural corpus order.
\end{abstract}

\section{Introduction}
It is well established that in language acquisition, there are robust patterns in the order by which phenomena are acquired.
For example, prototypical concepts are acquired earlier; concrete words tend to be learned before abstract ones \cite{rosch1978cognition}.
The acquisition of lexical knowledge in artificial systems proceeds differently. In general, models will improve during the course of parameter learning, but the time course of acquisition is not generally studied beyond generalization error as a function of training time or data size.
We revisit this issue of choosing the order of learning---\textbf{curriculum learning}---framing it as an optimization problem so that a rich array of factors---including nuanced measures of difficulty, as well as prototypicality and diversity---can be exploited.

Prior research focusing on curriculum strategies in NLP is scarce, and has conventionally been following a paradigm of ``starting small'' \cite{elman1993learning}, i.e., initializing the learner with ``simple'' examples first, and then gradually increasing data complexity \cite{bengio2009CL,spitkovsky2010baby}. In language modeling, this preference for increasing complexity has been realized by curricula that increase the entropy of training data by growing the size of the training vocabulary from frequent to less frequent words \cite{bengio2009CL}. In unsupervised grammar induction, an effective curriculum comes from increasing length of training sentences as training progresses \cite{spitkovsky2010baby}. These case studies have demonstrated that carefully designed curricula can lead to better results. However, they have relied on heuristics in selecting curricula or have followed the intuitions of human and animal learning \cite{kail1990development,skinner1938behavior}. Had different heuristics been chosen, the results would have been different. In this paper, we use curriculum learning to create improved word representations. However, rather than testing a small number of curricula, we search for an optimal curriculum using Bayesian optimization.
A curriculum is defined to be the ordering of the training instances,
in our case it is the ordering of paragraphs in which the representation learning  model reads the corpus.
We use a linear ranking function to conduct a systematic exploration of interacting factors that affect curricula of representation learning models. We then analyze our findings, and compare them to human intuitions and learning principles.

We treat curriculum learning as an outer loop in the process of learning and evaluation of vector-space representations of words; the iterative procedure is (1) predict a curriculum; (2) train word embeddings; (3) evaluate the embeddings on tasks that use word embeddings as the sole features. Through this model we analyze the impact of curriculum on word representation models and on extrinsic tasks.
To quantify curriculum properties, we define three groups of features aimed at analyzing statistical and linguistic content and structure of training data: (1) diversity, (2) simplicity, and (3) prototypicality. A function of these features is computed to score each paragraph in the training data, and the curriculum is determined by sorting corpus paragraphs by the paragraph scores. We detail the model in \Sref{sec:model}. Word vectors are learned from the sorted corpus, and then evaluated on part-of-speech tagging, parsing, named entity recognition, and sentiment analysis (\Sref{sec:tasks}). Our experiments confirm that training data curriculum affects model performance, and that models with optimized curriculum consistently outperform baselines trained on shuffled corpora (\Sref{sec:experiments}). We analyze our findings in \Sref{sec:analysis}.

The contributions of this work are twofold. First, this is the first framework that formulates curriculum learning as an optimization problem, rather then shuffling data or relying on human intuitions. We experiment with optimizing the curriculum of word embeddings, but in principle the curriculum of other models can be optimized in a similar way. Second, to the best of our knowledge, this study is the first to analyze the impact of distributional and linguistic properties of training texts on the quality of task-specific word embeddings.

\section{Curriculum Learning Model}
\label{sec:model}
We are considering the problem of maximizing a performance of an NLP task through sequentially optimizing the curriculum of training data of word vector representations that are used as features in the task.

Let $\mathcal{X}=\{x_1,x_2,\ldots,x_n\}$ be the training corpus with $n$ lines (sentences or paragraphs).
The curriculum of word representations is quantified by scoring each of the paragraphs according to the linear function $\mathbf{w}^{\intercal}\mathbf{\boldsymbol{\phi}}(\mathcal{X})$, 
where $\boldsymbol{\phi}(\mathcal{X}) \in \mathbb{R}^{\ell \times 1}$ is a real-valued vector containing  $\ell$ linguistic features extracted for each paragraph, and $\mathbf{w} \in \mathbb{R}^{\ell \times 1}$ denote the weights learned for these features. The feature values $\boldsymbol{\phi}(\mathcal{X})$ are $z$-normalized across all paragraphs.
These scores are used to specify the order of the paragraphs in the corpus---the curriculum: we sort the paragraphs by their scores.

After the paragraphs are curriculum-ordered, the reordered training corpus is used to generate word representations.
These word representations are then used as features in a subsequent NLP task.
We define the objective function $eval : \mathcal{X}\rightarrow\mathbb{R}$, which is the quality estimation metric for this NLP task performed on a held-out dataset (e.g., correlation, accuracy, $F_1$ score, BLEU).
Our goal is to define the features $\boldsymbol{\phi}(\mathcal{X})$ and to find the optimal weights $\mathbf{w}$ that maximize $eval$.

\begin{figure}[!ht]
  \centering
  \includegraphics[width=1.05\columnwidth]{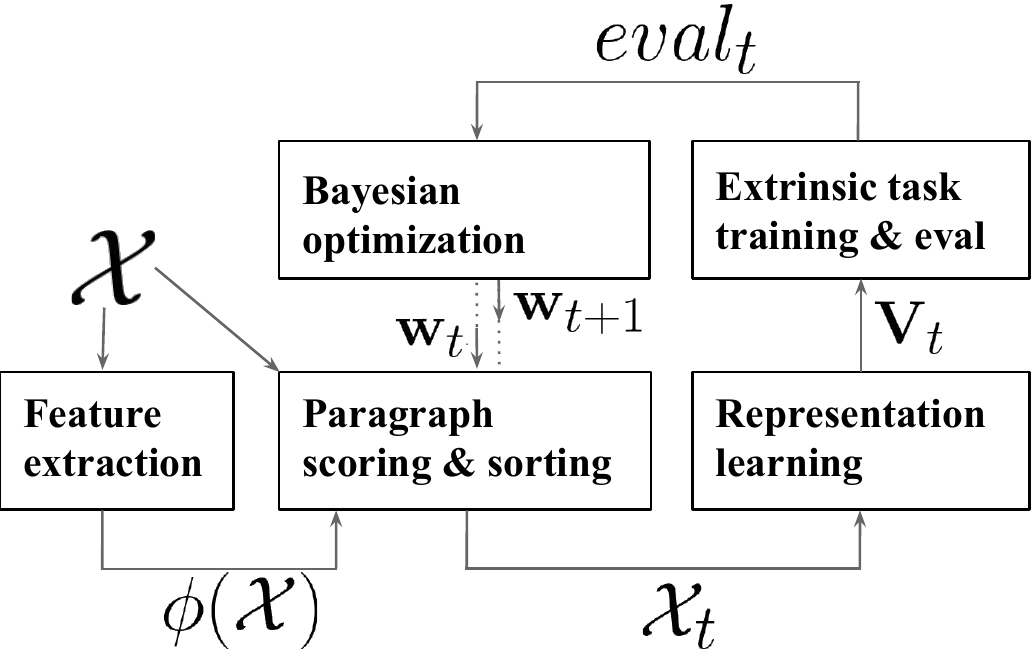}
  \caption{Curriculum optimization framework.}
  \label{fig:flow}
\end{figure}

We optimize the feature weights using Bayesian optimization; we detail the model in \Sref{sec:BO}.
Distributional and linguistic features inspired by prior research in language acquisition and second language learning are described in \Sref{sec:features}. \Fref{fig:flow} shows the computation flow diagram.

\subsection{Bayesian Optimization for Curriculum Learning}
\label{sec:BO}

As no assumptions are made regarding the form of $eval(\mathbf{w})$, gradient-based methods cannot be applied, and performing a grid search over parameterizations of $\mathbf{w}$ would require a exponentially growing number of parameterizations to be traversed. Thus, we propose to use Bayesian Optimization (BayesOpt) as the means to maximize $eval(\mathbf{w})$. BayesOpt is a methodology to globally optimize \textit{expensive}, multimodal black-box functions \cite{shahriari2016taking,bergstra2011bayesopt,snoek2012practical}.
It can be viewed as a sequential approach to performing a regression from high-level model parameters (e.g., learning rate, number of layers in a neural network, and in our model--curriculum weights $\mathbf{w}$) to the loss function or the performance measure ($eval$).


An arbitrary objective function, $eval$, is treated as a black-box, and BayesOpt uses Bayesian inference to characterize a posterior distribution over functions that approximate $eval$. This model of $eval$ is called the \textbf{surrogate model}. Then, the BayesOpt exploits this model to make decisions about $eval$, e.g., where is the expected maximum of the function, and what is the expected improvement that can be obtained over the best iteration so far. The strategy function, estimating the next set of parameters to explore given the current beliefs about $eval$ is called the \textbf{acquisition function}. The surrogate model and the acquisition function are the two key components in the BayesOpt framework; their interaction is shown in \Aref{algorithm}.

The surrogate model allows us to cheaply approximate the quality of a set of parameters $\mathbf{w}$ without running $eval(\mathbf{w})$, and the acquisition function uses this surrogate to choose a new value of $\mathbf{w}$. However, a trade-off must be made: should the acquisition function move $\mathbf{w}$ into a region where the surrogate believes an optimal value will be found, or should it explore regions of the space that reveal more about how $eval$ behaves, perhaps discovering even better values? That is, acquisition functions balance a tradeoff between exploration---by selecting $\mathbf{w}$ in the regions where the uncertainty of the surrogate model is high, and exploitation---by querying the regions where the model prediction is high. 

Popular choices for the surrogate model are Gaussian Processes \cite[GP]{rasmussen2006gaussian,snoek2012practical}, providing convenient and powerful prior distribution on functions, and tree-structured Parzen estimators \cite[TPE]{bergstra2011bayesopt}, tailored to handle conditional spaces.
Choices of the acquisition functions include probability of improvement \cite{kushner1964new}, expected improvement (EI) \cite{movckus1975bayesian,jones2001taxonomy}, GP upper confidence bound \cite{srinivas2009gaussian}, Thompson sampling \cite{thompson1933likelihood}, entropy search \cite{hennig2012entropy}, and dynamic combinations of the above functions \cite{hoffman2011portfolio}; see \newcite{shahriari2016taking} for an extensive comparison.
\newcite{dani2015bayesopt} found that the combination of EI as the acquisition function and TPE as the surrogate model performed favorably in Bayesian optimization of text representations; we follow this choice in our model.

\begin{algorithm}
\caption{Bayesian optimization}
\label{algorithm}
\begin{algorithmic}[1]
\State $\mathcal{H} \leftarrow  \emptyset$ \hfill \Comment Initialize observation history
\State $\mathcal{A} \leftarrow  EI$ \hfill \Comment Initialize acquisition function
\State $\mathcal{S}_0 \leftarrow  TPE$ \hfill \Comment Initialize surrogate model
\For{t $\leftarrow$ 1 to $T$ }
\State $\mathbf{w}_{t} \leftarrow \mathrm{argmax}_{\mathbf{w}}\mathcal{A}(\mathbf{w}; \mathcal{S}_{t-1}, \mathcal{H})$ \Comment Predict $\mathbf{w}_{t}$ by optimizing acquisition function
\State $eval(\mathbf{w}_{t})$  \hfill \Comment Evaluate $\mathbf{w}_{t}$ on extrinsic task
\State $\mathcal{H} \leftarrow  \mathcal{H}\cup(\mathbf{w}_{t},eval(\mathbf{w}_{t}))$ \Comment Update observation history
\State Estimate $\mathcal{S}_{t}$ given $\mathcal{H}$
\EndFor
\State \Return $\mathcal{H}$
\end{algorithmic}
\end{algorithm}

\subsection{Distributional and Linguistic Features}
\label{sec:features}
To characterize and quantify a curriculum, we define three categories of features, focusing on various distributional, syntactic, and semantic aspects of training data. We now detail the feature categories along with motivations for feature selection.
\paragraph{\textsc{diversity}.}
Diversity measures capture the distributions of types in data. Entropy is the best-known measure of diversity in statistical research, but there are many others \cite{tang2006analysis,gimpel13diversity}.
Common measures of diversity are used in many contrasting fields, from ecology and biology \cite{rosenzweig1995species,magurran2013measuring}, to economics and social studies \cite{stirling2007general}.
Diversity has been shown effective in related research on curriculum learning in language modeling, vision, and multimedia analysis \cite{bengio2009CL,jiang2014diversity}.

Let $p_i$ and $p_j$ correspond to empirical frequencies of word types $t_i$ and $t_j$ in the training data.
Let $d_{ij}$ correspond to their semantic similarity, calculated as the cosine similarity between embeddings of $t_i$ and $t_j$ learned from the training data.
We annotate each paragraph with the following diversity features:
\begin{itemizesquish}
\item Number of word types: $\#types$
\item Type-token ratio: $\frac{\#types}{\#tokens}$
\item Entropy: $- \sum _i p_i \mathrm{ln} (p_i)$ 
\item Simpson's index \cite{simpson1949measurement}: $\sum _i p{_i}{^2}$
\item Quadratic entropy \cite{rao1982diversity}:\footnote{Intuitively, this feature promotes paragraphs that contain semantically similar high-probability words.} $\sum _{i,j} d_{ij}p{_i}p{_j}$
\end{itemizesquish}


\paragraph{\textsc{simplicity}.}
\newcite{spitkovsky2010baby} have validated the utility of syntactic simplicity in curriculum learning for unsupervised grammar induction by showing that training on sentences in order of increasing lengths outperformed other orderings.
We explore the simplicity hypothesis, albeit without prior assumptions on specific ordering of data, and extend it to additional simplicity/complexity measures of training data.
Our features are inspired by prior research in second language acquisition, text simplification, and readability assessment \cite{Schwarm:2005,callan2007combining,pitler2008revisiting,vajjala2012improving}.
We use an off-the-shelf syntactic parser\footnote{\url{http://http://people.sutd.edu.sg/~yue_zhang/doc}} \cite{zpar} to parse our training corpus.
Then, the following features are used to measure phonological, lexical, and syntactic complexity of training paragraphs:
\begin{itemizesquish}
\item Language model score
\item Character language model score
\item Average sentence length
\item Verb-token ratio
\item Noun-token ratio
\item Parse tree depth
\item Number of noun phrases: $\#NPs$
\item Number of verb phrases: $\#VBs$
\item Number of prepositional phrases: $\#PPs$
\end{itemizesquish}

\paragraph{\textsc{prototypicality}.}
This is a group of semantic features that use insights from cognitive linguistics and child language acquisition.
The goal is to characterize the curriculum of representation learning in terms of the curriculum of human language learning. 
We resort to the Prototype theory \cite{rosch1978cognition}, which posits that
semantic categories include more central (or prototypical) as well as less prototypical words.
For example, in the \textsc{animal} category, \textit{dog} is more prototypical than \textit{sloth} (because \textit{dog} is more frequent); \textit{dog} is more prototypical than \textit{canine} (because \textit{dog} is more concrete); and \textit{dog} is more prototypical than \textit{bull terrier} (because \textit{dog} is less specific).
According to the theory, more prototypical words are acquired earlier.
We use lexical semantic databases to operationalize insights from the prototype theory in the following semantic features; the features are computed on token level and averaged over paragraphs:
\begin{itemizesquish}
\item Age of acquisition (AoA) of words was extracted from the crowd-sourced database, containing over 50 thousand English words \cite{kuperman2012age}. For example, the AoA of \textit{run} is 4.47 (years), of \textit{flee} is 8.33, and of \textit{abscond} is 13.36. If a word was not found in the database it was assigned the maximal age of 25.
\item Concreteness ratings on the scale of 1--5 (1 is most abstract) for  40 thousand English lemmas \cite{brysbaert2014concreteness}.  For example, \textit{cookie} is rated as 5, and \textit{spirituality} as 1.07.
\item Imageability ratings are taken from the MRC psycholinguistic database \cite{Wilson87mrcpsycholinguistic}. Following \newcite{tsvetkov14metaphor}, we used the MRC annotations as seed, and propagated the ratings to all vocabulary words using the word embeddings as features in an $\ell_2$-regularized logistic regression classifier.
\item Conventionalization features count the number of ``conventional'' words and phrases in a paragraph. Assuming that a Wikipedia title is a proxy to a conventionalized concept, we counted the number of existing titles (from a database of over 4.5 million titles) in the paragraph. 
\item Number of syllables scores are also extracted from the AoA database; out-of-database words were annotated as 5-syllable words.
\item Relative frequency in a supersense was computed by marginalizing the word frequencies in the training corpus over coarse semantic categories defined in the WordNet \cite{wordnet,ciaramita2006}. There are 41 supersense types: 26 for nouns and 15 for verbs, e.g., \textsc{noun.animal} and \textsc{verb.motion}. For example, in \textsc{noun.animal} the relative frequency of \textit{human} is 0.06, of \textit{dog} is 0.01, of \textit{bird} is 0.01, of \textit{cattle} is 0.009, and of \textit{bumblebee} is 0.0002.
\item Relative frequency in a synset was calculated similarly to the previous feature category, but word frequencies were marginalized over WordNet synsets (more fine-grained synonym sets). For example, in the synset \{\textit{vet, warhorse, veteran, oldtimer, seasoned stager}\}, \textit{veteran} is the most prototypical word, scoring 0.87.
\end{itemizesquish}

\section{Evaluation Benchmarks}
\label{sec:tasks}
We evaluate the utility of the pretrained word embeddings as features
in downstream NLP tasks. We choose the following off-the-shelf models that utilize
pretrained word embeddings as features:

\paragraph{Sentiment Analysis (Senti).}
\newcite{socher2013senti} created a treebank of sentences
annotated with fine-grained sentiment labels on phrases and sentences
from movie review excerpts.
The coarse-grained treebank of positive and negative
classes has been split into training, development, and test datasets
containing 6,920, 872, and 1,821 sentences, respectively.
We use the average of the word vectors of a given sentence as a feature vector for
classification \cite{faruqui:2015:Retro,sedoc2016semantic}.
The $\ell_2$-regularized logistic regression classifier is
tuned on the development set and accuracy is reported on the test set.

\paragraph{Named Entity Recognition (NER).}
Named entity recognition is the task of identifying proper names in a sentence, such as names of persons, locations etc. We use the recently proposed LSTM-CRF NER model \cite{lample2016neural} which trains a forward-backward LSTM on a given sequence of words (represented as word vectors), the hidden units of which are then used as (the only) features in a CRF model \cite{lafferty2001conditional} to predict the output label sequence. We use the CoNLL 2003 English NER dataset \cite{tjong2003introduction} to train our models and present results on the test set.

\paragraph{Part of Speech Tagging (POS).}
For POS tagging, we again use the LSTM-CRF model \cite{lample2016neural}, but
instead of predicting the named entity tag for every word in a sentence, we
train the tagger to predict the POS tag of the word.
The tagger is trained and evaluated with the standard Penn TreeBank (PTB)
\cite{marcus1993building} training, development and test set splits as
described in \newcite{collins2002discriminative}.

\paragraph{Dependency Parsing (Parse).}
Dependency parsing is the task of identifying syntactic relations between the
words of a sentence. For dependency parsing, we train the stack-LSTM
parser of \newcite{dyer2015transition} for English on the universal
dependencies v1.1 treebank \cite{agic:2015} with the standard development and test splits,
reporting unlabeled attachment scores (UAS) on the test data.
We remove all part-of-speech and morphology features from the
data, and prevent the model from optimizing the word embeddings used
to represent each word in the corpus, thereby forcing
the parser to rely completely on the pretrained embeddings.

\section{Experiments}
\label{sec:experiments}

\paragraph{Data.}
All models were trained on 	Wikipedia articles, split to paragraph-per-line. Texts were cleaned, tokenized, numbers were normalized by replacing each digit with ``DG'', all types that occur less than 10 times were replaces by the ``UNK'' token, the data was not lowercased. We list data sizes in \tref{tbl:data-sizes}.

\begin{table}[h]
\centering
\begin{tabular}{ccc}
\# paragraphs  & \# tokens & \# types\\\hline
2,532,361  & 100,872,713 & 156,663\\
\end{tabular}
\caption{Training data sizes.}
\label{tbl:data-sizes}
\end{table}

\paragraph{Setup.}
100-dimensional word embeddings were trained using the \texttt{cbow} model implemented in the word2vec toolkit~\cite{mikolov-iclr13}.\footnote{To evaluate the impact of curriculum learning, we enforced sequential processing of data organized in a pre-defined order of training examples. To control for sequential processing, word embedding were learned by running the \texttt{cbow} using a single thread for one iteration.}
All training data was used, either shuffled or ordered by a curriculum.
As described in \Sref{sec:tasks}, we modified the extrinsic tasks to learn solely from word embeddings, without additional features.
All models were learned under same conditions, across curricula: in Parse, NER, and POS  we limited the number of training iterations to 3, 3, and 1, respectively. This setup allowed us to evaluate the effect of curriculum without additional interacting factors. 

\paragraph{Experiments.}
In all the experiments we first train word embedding models, then the word embeddings are used as features in four extrinsic tasks (\Sref{sec:tasks}).  We tune the tasks on development data, and report results on the test data.
The only component that varies across the experiments is order of paragraphs in the training corpus---the curriculum. We compare the following experimental setups:
\begin{itemizesquish}
\item \textbf{Shuffled} baselines: the curriculum is defined by random shuffling the training data. We shuffled the data 10 times, and trained 10 word embeddings models, each model was then evaluated on downstream tasks. Following \newcite{bengio2009CL}, we report test results for the system that is closest to the median in dev scores. To evaluate variability and a range of scores that can be obtained from shuffling the data, we also report test results for systems that obtained the highest dev scores.
\item \textbf{Sorted} baselines: the curriculum is defined by sorting the training data by sentence length in increasing/decreasing order, similarly to \cite{spitkovsky2010baby}. 
\item \textbf{Coherent} baselines: the curriculum is defined by just concatenating Wikipedia articles. The goal of this experiment is to evaluate the importance of semantic coherence in training data. Our intuition is that a coherent curriculum can improve models, since words with similar meanings and similar contexts are grouped when presented to the learner.
\item \textbf{Optimized curriculum} models: the curriculum is optimized using the BayesOpt.
We evaluate and compare models optimized using features from one of the three feature groups (\Sref{sec:features}).
As in the shuffled baselines, we fix the number of trials (here, BayesOpt iterations) to 10, and we report test results of systems that obtained best dev scores.
\end{itemizesquish}

\paragraph{Results.}
Experimental results are listed in \tref{tbl:results}.
Most systems trained with curriculum substantially outperform the  strongest of all baselines.
These results are encouraging, given that all word embedding models were trained on the same set of examples,
only in different order, and display the indirect influence of the data curriculum on downstream tasks.
These results support our assumption that curriculum matters.
Albeit not as pronounced as with optimized curriculum, sorting paragraphs by length
can also lead to substantial improvements over random baselines, but there is no clear recipe on whether the models prefer curricula sorted in an increasing or decreasing order.
These results also support the advantage of a task-specific optimization framework over a general, intuition-guided recipe.
An interesting result, also, that shuffling is not essential: systems trained on coherent data are on par (or better) than the shuffled systems.\footnote{Note that in the shuffled NER baselines, best dev results yield lower performance on the test data.
This implies that in the standard development/test splits the development and test sets are not fully compatible or not large enough.
We also observe this problem in the curriculum-optimized Parse-prototypicality and Senti-diversity systems.
The dev scores for the Parse systems are 76.99, 76.47, 76.47 for diversity, prototypicality, and simplicity, respectively, but the prototypicality-sorted parser performs poorly on test data. Similarly in the sentiment analysis task, the dev scores are 69.15, 69.04, 69.49 for diversity, prototypicality, and simplicity feature groups.    Senti-diversity scores, however, are lower on the test data, although the dev results are better than in Senti-simplicity.
This limitation of the standard dev/test splits is beyond the scope of this paper. }
In the next section, we analyze these results qualitatively.
\begin{table*}[ht!]
\centering
\begin{tabular}{clc|c|c|c}
& &\multicolumn{1}{c}{\textbf{Senti}} & \multicolumn{1}{c}{\textbf{NER}}  & \multicolumn{1}{c}{\textbf{POS}} & \multicolumn{1}{c}{\textbf{Parse}} \\
\cline{2-6}
\multirow{2}{*}{\textbf{Shuffled}} 
									& median & 66.01 & 85.88 & 96.35  & 75.08 \\
                                   & best   & 66.61 & 85.50 & 96.38  & 76.40 \\
\cline{2-6}
\multirow{2}{*}{\textbf{Sorted}} &	long$\rightarrow$short & 66.78 & 85.22 & 96.47  & 75.85 \\
                                 &	short$\rightarrow$long & 66.12 & 85.49 & 96.20  & 75.31 \\
\cline{2-6}
\textbf{Coherent}				& original order & 66.23 & 85.99 & 96.47  & 76.08 \\
\cline{2-6}
\multirow{3}{*}{\textbf{\begin{tabular}[c]{@{}c@{}}Optimized \\ curriculum\end{tabular}}}
								& diversity       & 66.06 & 86.09 & 96.59 & \textbf{76.63} \\
                               & prototypicality & \textbf{67.44} & 85.96 & 96.53 & 75.81 \\
                               & simplicity      & 67.11 & \textbf{86.42} & \textbf{96.62} & 76.54 \\
\end{tabular}
\caption{Evaluation of the impact of the curriculum of word embeddings on the downstream tasks. }
\label{tbl:results}
\end{table*}

\section{Analysis}
\label{sec:analysis}
\paragraph{What are task-specific curriculum preferences?}
We manually inspect learned features and curriculum-sorted corpora, and find that best systems
are obtained when their embeddings are learned from curricula appropriate to the downstream tasks.
We discuss below several examples.

POS and Parse systems converge to the same set of weights, when trained on features that provide various measures of syntactic simplicity. The features with highest coefficients (and thus the most important features in sorting) are  $\#NPs$, Parse tree depth, $\#VPs$, and $\#PPs$ (in this order).
The sign in the $\#NPs$ feature weight, however, is the opposite from the other three feature weights
(i.e., sorted in different order). $\#NPs$ is sorted in the increasing order of the number of noun phrases in a paragraph,
and the other features are sorted in the decreasing order.
Since Wikipedia corpus contains a lot of partial phrases (titles and headings), such curriculum promotes more complex, full sentences, and demotes partial sentences.

Best Senti system is sorted by prototypicality features.
Most important features (with the highest coefficients) are Concreteness, Relative frequency in a supersense, and the Number of syllables. First two are sorted in decreasing order (i.e. paragraphs are sorted from more to less concrete, and from more to less prototypical words), and the Number of syllables is sorted in increasing order (this also promotes simpler, shorter words which are more prototypical).
We hypothesize that this soring reflects the type of data that Sentiment analysis task is trained on: it is trained on movie reviews, that are usually written in a simple, colloquial language.

\begin{figure}[!ht]
  \centering
  \includegraphics[width=0.8\columnwidth]{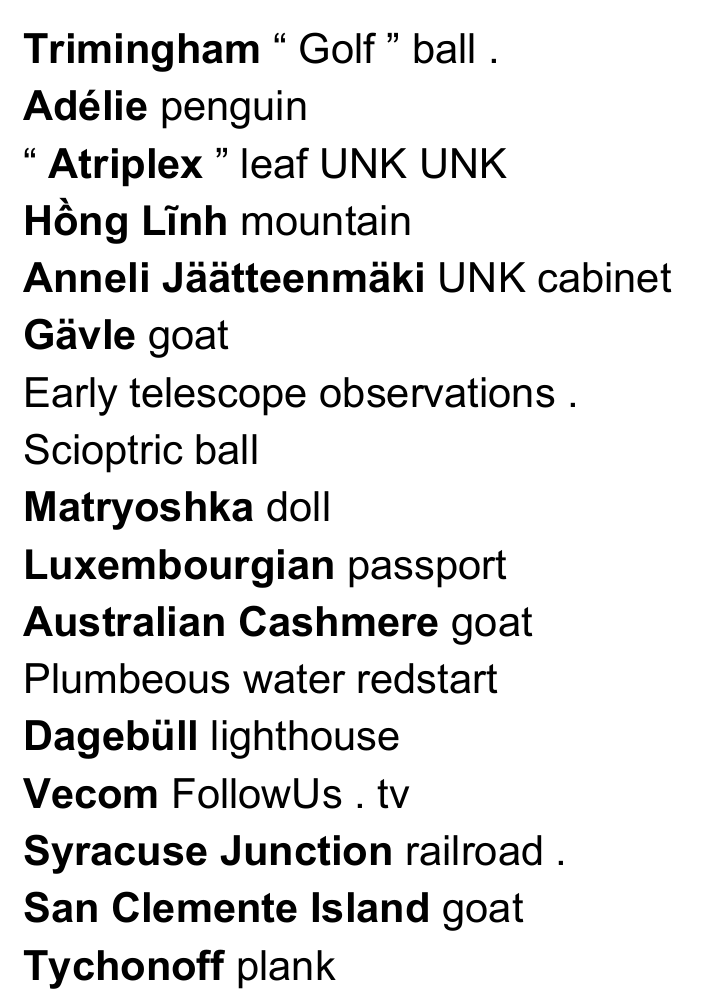}
  \caption{Most of the top lines in best-scoring NER system contain named entities, although our features do not annotate named entities explicitly.}
  \label{fig:ner_simplicity}
\end{figure}

Unlike POS, Parse, and Senti systems, all NER systems prefer curricula in which texts are sorted from short to long paragraphs. The most important features in the best (simplicity-sorted) system are  $\#PPs$ and Verb-token ratio, both sorted from less to more occurrences of prepositional and verb phrases. Interestingly, most of the top lines in the NER system curricula contain named entities, although none of our features mark named entities explicitly.
We show top lines in the simplicity-optimized system in \fref{fig:ner_simplicity}.

Finally, in all systems sorted by prototypicality, the last line is indeed not a prototypical word \textit{Donaudampfschiffahrtselektrizit{\"a}tenhauptbetriebswerkbauunterbeamtengesellschaft}, which is an actual word in German, frequently used as an example of compounding in synthetic languages, but rarely (or never?) used by German speakers.

\paragraph{Weighting examples according to curriculum.}
Another way to integrate curriculum in word embedding training is to weight training examples according to curriculum during word representation training.
We modify the \texttt{cbow} objective $\sum_{t=1}^T \log p(w_t | w_{t-c} .. w_{t+c})$ as follows:\footnote{The modified word2vec tool is located at \url{https://github.com/wlin12/wang2vec} .}
\begin{equation*}
\sum_{t=1}^T (\frac{1}{1+e^{-weight(w_t)}}+\lambda) \log p(w_t | w_{t-c} .. w_{t+c})
\end{equation*}
Here, $weight(w_t)$ denotes the score attributed to the token $w_t$, which is the $z$-normalized score of the paragraph; $\lambda$=0.5 is determined empirically.
$\log p(w_t)|w_{t-c} .. w_{t+c})$ computes the probability of predicting word $w_t$, using the context of $c$ words to the left and right of $w_t$.
Notice that this quantity is no longer a proper probability, as we are not normalizing over the weights $weight(w_t)$ over all tokens. However, the optimization in word2vec is performed using stochastic gradient descent, optimizing for a single token at each iteration. This yields a normalizer of 1 for each iteration, yielding the same gradient as the original \texttt{cbow} model.

We retrain our best curriculum-sorted systems with the modified objective, also controlling for curriculum. The results are shown in \tref{tbl:w2v-objective}.
We find that the benefit of integrating curriculum in training objective of word representations is not evident across tasks:
Senti and NER systems trained on vectors with the modified objective substantially outperform best results in \tref{tbl:results}; POS and Parse perform better than the baselines but worse than the systems with the original objective.
\begin{table}[h]
\centering
\begin{tabular}{l|rrrr}
              & \textbf{Senti} & \textbf{NER} & \textbf{POS} & \textbf{Parse} \\\hline
curriculum		& 67.44 & 86.42 & \textbf{96.62}  & \textbf{76.63} \\
\texttt{cbow}+curric		& \textbf{68.26} & \textbf{86.49} & 96.48 & 76.54 \\
\end{tabular}
\caption{Evaluation of the impact of curriculum integrated in the \texttt{cbow} objective.}
\label{tbl:w2v-objective}
\end{table}

\paragraph{Are we learning task-specific curricula?}
One way to assess whether we learn meaningful task-specific curriculum preferences is to compare
curricula learned by one downstream task across different feature groups.
If learned curricula are similar in, say, NER system, despite being optimized once using diversity features and once using prototypicality features---two disjoint feature sets---we can infer that the NER task prefers word embeddings learned from examples presented in a certain order, regardless of specific optimization features.
For each downstream task, we thus measure Spearman's rank correlation between the curricula optimized using diversity (\textsc{d}), or prototypicality (\textsc{p}), or simplicity (\textsc{s}) feature sets. Prior to measuring correlations, we remove duplicate lines from the training corpora.
Correlation results across tasks and across feature sets are shown in \tref{tbl:rho}.

The general pattern of results is that if two systems score higher than baselines,
training sentences of their feature embeddings have similar curricula (i.e., the Spearman's $\rho$ is positive),
and if two systems disagree (one is above and one is below the baseline),
then their curricula also disagree (i.e., the Spearman's $\rho$ is negative or close to zero).
NER systems all outperform the baselines and their curricula have high correlations.
Moreover, NER sorted by diversity and simplicity have better scores than NER sorted by prototypicality,
and in line with these results $\rho$(\textsc{s,d})$_{NER} > \rho$(\textsc{p,s})$_{NER}$ and $\rho$(\textsc{s,d})$_{NER} > \rho$(\textsc{d,p})$_{NER}$.
Similar pattern of results is in POS correlations.
In Parse systems, also, diversity and simplicity features yielded best parsing results, and
$\rho$(\textsc{s,d})$_{Parse}$ has high positive correlation.
The prototypicality-optimized parser performed poorly, and its correlations with better systems are negative.
The best parser was trained using the diversity-optimized curriculum,
and thus $\rho$(\textsc{d,p})$_{Parse}$ is the lowest. Senti results follow similar pattern of curricula correlations.
\begin{table}[ht]
\centering
\begin{tabular}{l|rrrr}
              & \textbf{Senti} & \textbf{NER} & \textbf{POS} & \textbf{Parse} \\\hline
$\rho$(\textsc{d, p}) & -0.68          & 0.76         & 0.66         &       -0.76 \\
$\rho$(\textsc{p, s}) & 0.33           & 0.75         & 0.75         &       -0.45 \\
$\rho$(\textsc{s, d}) & -0.16          & 0.81         & 0.51         &        0.67 \\
\end{tabular}
\caption{Curricula correlations across feature groups.}
\label{tbl:rho}
\end{table}

\paragraph{Curriculum learning vs. data selection.}
We compare the task of curriculum learning to the task of data selection (reducing the set of training instances to more important or cleaner examples).
We reduce the training data to the subset of 10\% of tokens, and train downstream tasks on the reduced training sets.
We compare system performance trained using the top 10\% of tokens in the best curriculum-sorted systems (Senti-prototypicality, NER-implicity, POS-simplicity, Parse-diversity) to the systems trained using the top 10\% of tokens in a corpus with randomly shuffled paragraphs.\footnote{Top $n$\% tokens are used rather than top $n$\% paragraphs because in all tasks except NER curriculum-sorted corpora begin with longer paragraphs. Thus, with top $n$\% paragraphs our systems would have an advantage over random systems due to larger vocabulary sizes and not necessarily due to a better subset of data.}
The results are listed in \tref{tbl:data-selection}.


\begin{table}[h]
\centering
\begin{tabular}{l|rrrr}
              & \textbf{Senti} & \textbf{NER} & \textbf{POS} & \textbf{Parse} \\\hline
random		& 63.97 			& \textbf{82.35} & 96.22 & 69.11 \\
curriculum	& \textbf{64.47} & 76.96 & \textbf{96.55} & \textbf{72.93} \\
\end{tabular}
\caption{Data selection results.}
\label{tbl:data-selection}
\end{table}

The curriculum-based systems are better in POS and in Parse systems, mainly because these tasks prefer vectors trained on curricula that promote well-formed sentences (as discussed above).
Conversely, NER prefers vectors trained on corpora that begin with named entities, so most of the tokens in the reduced training data are constituents in short noun phrases.
These results suggest that the tasks of data selection and curriculum learning are different.
Curriculum is about strong initialization of the models and time-course learning, which is not necessarily sufficient for data reduction.

\section{Related Work}
Two prior studies on curriculum learning in NLP are discussed in the paper \cite{bengio2009CL,spitkovsky2010baby}.
Curriculum learning and related research on self-paced learning has been explored more deeply in computer vision \cite{bengio2009CL,NIPS2010_3923,lee2011learning} and in multimedia analysis \cite{jiang2015self}.
Bayesian optimization has also received little attention in NLP.  GPs were used in the task of machine translation quality estimation \cite{cohn2013modelling} and in temporal analysis of social media texts \cite{preotiuc2013temporal};
TPEs were used by \newcite{dani2015bayesopt} for optimizing choices of feature representations---$n$-gram size, regularization choice, etc.---in supervised classifiers.

\section{Conclusion}
We used Bayesian optimization to optimize curricula for training dense distributed word representations, which, in turn, were used as the sole features in NLP tasks. Our experiments confirmed that better curricula yield stronger models. We also conducted an extensive analysis, which sheds better light on understanding of text properties that are beneficial for model initialization. The proposed novel technique for finding an optimal curriculum is general, and can be used with other datasets and models.

\section*{Acknowledgments}
This work was supported by the National Science Foundation through award IIS-1526745.
We are grateful to Nathan Schneider, Guillaume Lample, Waleed Ammar, Austin Matthews, and the anonymous reviewers for their insightful comments.

\bibliography{curric}
\bibliographystyle{acl2016}
\end{document}